# Coordination and Machine Learning in Multi-Robot Systems: Applications in Robotic Soccer

**Luís Paulo Reis**

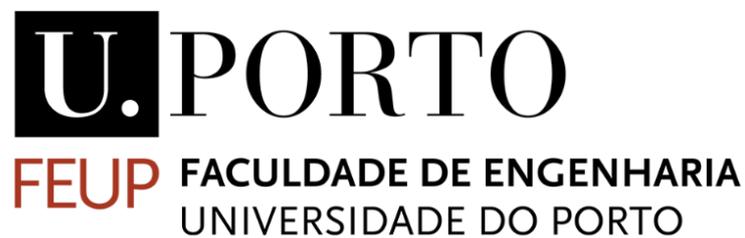

Porto, August 2022

# Table of Contents

# Abstract

This paper presents in a simple and very concise manner, and accessible to a broad audience, the concepts of Artificial Intelligence (AI), Agents and Multi-Agent Systems (MAS), Coordination in MAS, Intelligent Robotics (IR), Multi-Robot Systems (MRS), Machine Learning (ML), Reinforcement Learning (RL) and Deep Reinforcement Learning (DRL). Emphasis is given on and how AI and ML, with special focus on DRL, may be efficiently used to create efficient robot skills and coordinated robotic teams, capable of performing very complex actions and tasks, as a team, such as playing a game of soccer.

The paper also presents the concept of robotic soccer and the vision and structure of the RoboCup initiative and the world robotic soccer championship with emphasis on the Humanoid Simulation 3D league and the new challenges for the AI community that RoboCup, and in particular this competition, poses. The final topics presented at the paper are based on the research developed or coordinated by the author throughout the last 22 years on the areas of multi-agent systems and intelligent robotics with emphasis for its applications in the domain of robotic soccer in the context of the FC Portugal project.

Topics concerning coordination and machine learning in robotic soccer are presented, including several coordination methodologies developed, such as: Strategy, Tactics, Formations, Setplays, and Coaching Languages and the use of Machine Learning to optimize the use of this concepts. The topics presented also include novel stochastic search algorithms for black box optimization and their use in the optimization of omnidirectional walking skills, robotic multi-agent learning and the creation of a humanoid kick with controlled distance. Finally, new applications using variations of the Proximal Policy Optimization (PPO) algorithm and advanced modelling for robot and multi-robot learning are briefly explained with emphasis for our new humanoid sprinting and running skills and the newly developed and amazing humanoid robot soccer dribbling skill.

The paper also discusses in a more high-level manner the research developed on coordination and machine learning for robotic soccer by myself and my research group. Our research group focused on developing scientific work, sometimes sacrificing wining competitions (as other research group do by just tunning, hacking and developing simple algorithms and strategies that work on simple and very specific problems) but this always seemed the correct strategy for us. However, even with this approach, the team won several competitions in different leagues and mostly won many scientific awards at RoboCup. In total, our team won more than 40 awards in international competitions including a clear victory at the Humanoid Simulation 3D League at the last RoboCup 2022 competition (Bangkok, Thailand, July 11-17, 2022). In this competition we scored a total of 84 goals conceding only 2, winning all the games except for a draw in the preparation/seeding round. FC Portugal also won all the other challenges, including the Proxy challenge and the Free/Scientific Challenge at RoboCup 2022. In what concerns scientific publications, the FC Portugal project and the works briefly described in this paper enabled us to publish more than 100 papers in indexed international conferences and journals.

# 1. Introduction

One of the main goals of the research on Multi-Agent Systems (MAS) or a Multi-Robot Systems (MRS) is to develop teamwork and coordination strategies, in particular in dynamic, continuous, stochastic, partially observable and multi-agent/multi-robot environments such as robotic soccer. The coordination and teamwork strategies in these systems can be built using different types of models that allow agents to reason about the best individual action to perform, but also the actions that maximizes the team global performance. Teamwork strategies and machine learning methodologies are decisive to make use of predefined plans to coordinate the execution of cooperative actions in a team and to be able to create competitive robot ad multi robot skills for such complex domains.

In recent years, the development of machine learning algorithms, with emphasis for deep reinforcement learning algorithms made it impossible for teams to compete using simple handmade/tunned skills when compared with robust machine learning approaches to learn the individual and collective skills. Thus, this paper is based on explaining how to create individual and multi-robot skills for multi-robot teams using advanced machine learning algorithms and how to use them and coordinate a complex multi-robot team for developing a very complex task such as playing a game of soccer.

This paper is organized by first introducing, in a simple and concise manner, the concepts of Artificial Intelligence (AI), Agents and Multi-Agent Systems (MAS), Coordination in MAS, Intelligent Robotics (IR), Multi-Robot Systems (MRS), Machine Learning (ML), Reinforcement Learning (RL), Deep Reinforcement Learning (DRL) and how AI and ML may be efficiently used to create efficient robot skills and coordinated robotic teams, capable of performing very complex actions and tasks, as a team, such as playing a game of soccer.

The paper is organized in the following sections: Artificial Intelligence and Multi-Agent Systems; Machine Learning and Deep Reinforcement Learning; Intelligent Robotics and Robot Learning; RoboCup and Robotic Soccer; Coordination and Machine Learning in Robotic Soccer; Conclusions.

The paper starts with a brief and straightforward introduction to the concepts of Artificial Intelligence and Multi-Agent Systems and the analysis of some open challenges on these domains. Then, it follows an explanation of the base concepts concerning Machine Learning and Deep Learning followed by a more in-depth analysis of Deep Reinforcement Learning, including some recent advances on the research on this area and some still open challenges, mostly in what concerns its application for robotics and multi-robot systems. Afterwards, the concept of Intelligent Robotics is described together with some recent advances on this area followed by the challenges posed by multi-robot systems and its coordination. Robot Learning and Multi-Robot Learning are then analysed in more detail. A description of the RoboCup international competition and its leagues with emphasis on the Humanoid Simulation 3D league is then displayed in order to show the domain in which most of the work presented in the next section was applied. The final section includes a description of our research work on the area of coordination for multi-robot systems in RoboCup and Machine Learning with applications in RoboCup with emphasis for the author's team research results in the last 22 years.

## 2. Artificial Intelligence and Multi-Agent Systems

### 2.1 Artificial Intelligence

Intelligence may be seen as the capacity to solve new problems through the use of knowledge. In this context, Artificial Intelligence may be defined as the Science concerned with building intelligent machines, that is, machines that perform tasks that when performed by humans require intelligence [Russel and Norvig, 1995].

Weak Artificial Intelligence, also known as narrow AI is artificial intelligence that is focused on one single narrow task. During the last few years significant efforts were develop for creating also Strong Artificial Intelligence. Strong AI or Artificial General Intelligence (AGI) is the intelligence of a machine that could successfully perform any intellectual task that a human being can! Although still science fiction, significant advances have been made in the last years on this concept such as for example the AlphaZero that is able to learn to play different types of board games from scratch surpassing the best human experts [Silver et al, 2018].

There are several areas in AI such as: Knowledge Representation and Reasoning; Expert Systems; Problem Solving; Planning and Scheduling; Machine Vision; Agents and Multi-Agent Systems; Machine Learning; Natural Language Processing; Intelligent Robotics. In the last years huge advances were seen in the areas of Machine Learning, Natural Language Processing and Intelligent Robotics.

Despite its short life, AI has evolved considerably since its creation. This progression was supported in part by major hardware advances, including processing speed, parallel computation architectures, storage capabilities, cost and accessibility to the general public. Also new machine learning frameworks and new algorithms are constantly being researched to minimize computational cost and maximize performance. This phenomenon has changed society and the way we look at data, which is now one of the most valuables assets a company can have.

The number of AI applications grew significantly in areas such as Health, Digital Media, Environment, Public Administration, among many other. For example, in the area of Health applications emerged in Radiology (X-ray, CT, MRI), Dermatology (Image analysis), Drug/Treatment Discovery, Risk Identification in Patients, Primary Care and Screening, Health Monitoring/Wearables, Cognitive and Social Rehabilitation, Physical Rehabilitation, Patient Interaction with the Health System, Health Systems Exchange of Information, Surgical/Medical Robots and Efficient Resource Allocation in the health system.

In Digital media, the AI applications become huge in the last few years ranging from Digital Images, Digital Video, Digital Audio, Movie Industry, Video Games, Web Pages and Websites, Social Media, Digital Data and Databases, Electronic Documents/Books. For example, in the area of Video Games AI now is crucial for developing engaging video-games with applications such as generating responsive, adaptive or intelligent behaviours, non-player characters with human-like intelligence, improve the game-player experience, game balancing and dynamic difficulty adjustment, movement patterns, in-game events based on player's input, pathfinding and decision trees for NPCs actions, procedural-content generation, text to speech and speech recognition, automatic level generation, AI opponents for board/strategic games among many others.

This work is mainly focused on the areas of Intelligent Robotics, Multi-Agent Systems and how Machine Learning may be applied for creating advanced robots and together with Coordination methodologies to create Multi-Robot Systems and Robotic Teams.

## 2.2 Agents and Multi-Agent Systems

An agent may be defined as a "Computational System, situated in a given environment, which has the ability to perceive that environment using sensors and act, in an autonomous way, in that environment using its actuators to fulfil a given function." [Russel and Norvig, 1995].

In this context a Multi-Agent System is a system that includes agents that for one side exhibit autonomous behaviour and on the other side interact with other agents in the system [Wooldridge, 1999].

There are several motivations for the use of multi-agent systems such as: handling problems of large dimensions; enabling the connection of legacy systems; being a natural solution for geographically or functionally distributed problems; the fact that knowledge or information are distributed; for project clarity and simplicity; for parallelization, speedup, and efficiency, particularly in tasks that can be decomposed into several subtasks, which are then handled asynchronously by the agents; fault tolerance and robustness, since the system suffers a gradual degradation as agents fail, instead of completely stopping; scalability, since more agents can be added to the MAS, in order to increase the parallelization of the system; problem division in smaller parts and maintenance of information privacy.

The inherent complexity of problems in MAS makes hand-tuned solutions very difficult to achieve and machine learning solutions start becoming a standard way of developing MAS and a new field of research emerged for Multi-Agent Learning with emphasis for Multi-Agent Deep Reinforcement Learning. One of the main challenges in this area is the coordination of a multi agent system. In fact, the motivation is that Agents do not live alone and have to work in a group cooperating and/or competing with other agents or humans.

## 2.3 Coordination in Multi-Agent/Multi-Robot Systems

Coordination may be defined as to work in harmony in a group. Coordination is needed due to the dependencies in agent actions; global constraints; the fact that typically agents individually do not have enough resources, information, or capacity to execute the task or solve the complete problem; efficiency by exchanging information or tasks division; and to prevent anarchy and chaos due to partial vision, lack of authority, conflicts, and agent's interactions.

Coordination is considered a key characteristic of MAS, and an agent's capability of coordinating with others constitutes one of its major qualities [Ossowski, 2008]. In cooperative environments, coordination consists of harmonizing the interactions of multiple agents, such that a global plan can be carried out. The global plan, possibly composed of the sum of each agent's individual actions, will ideally fulfil the agent's individual goals or the global objective of the MAS, as efficiently as possible. In competitive environments, coordination consists of finding the best strategies that complete an agent's own goals, while also avoiding being exploited by adversaries. Eventually, agents may converge to an equilibrium state where changing their policies will allow others to take advantage of them and eventually worsen their performance.

Flexible, high-level coordination and communication mechanisms are the key to deal with the dynamism, uncertainty and complexity of complex environments and tasks. To deploy and coordinate teams in these types of complex environments and tasks it is necessary to evaluate the team's performance and be able to dynamically reorganize the team according to its performance and contingencies. However, most implemented Multi-Agent Systems are not able to provide this flexibility in the coordination mechanisms. On the other hand, most of the methodologies for coordinating cooperative agents proposed in the literature are not concerned with the spatial mobility of agents and, as such, are not applicable to cooperative tasks performed by teams of robotic agents in real-world or real-world environments or simulations of these.

# 3. Machine Learning

## 3.1 Introduction

Arthur Samuel in 1959 defined Machine learning as "Machine Learning enables a machine to automatically learn from data, improve performance from experiences, and predict things without being explicitly programmed" [Samuel, 1959]. In a more complete definition "Machine learning is a field of artificial intelligence that gives computer systems the ability to "learn" (e.g., progressively improve performance on a specific task) from data/results of their actions, without being explicitly programmed".

Machine learning is changing the way programming is performed (Figure 1). In fact, traditional programming in which a program was developed based on very specific rules and algorithms and using its input was able to generate the desired output, is being replaced by machine learning. In this, machine learning new way of programming, data and output are used to feed a machine learning process that is able to create the program/model that is able to capture the concept and generate the new desired output from new inputs.

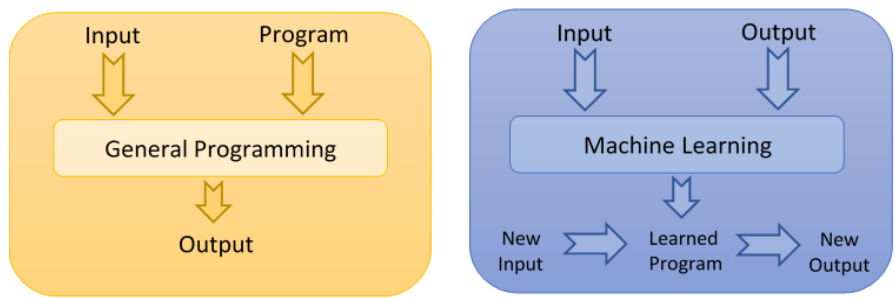

*Figure 1: General Programming vs Machine Learning*

At a broad level Machine Learning may be subdivided in three types: Supervised Learning (Classification and Regression), Unsupervised Learning (Clustering and Dimensionality Reduction) and Reinforcement Learning (Figure 2).

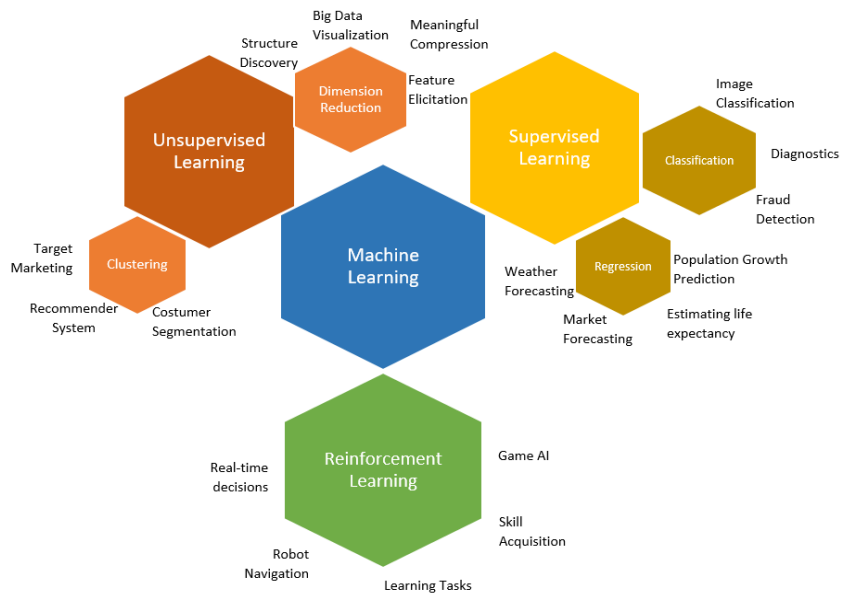

*Figure2: Machine Learning Types*

In supervised learning, we have labelled information with an output variable. We use a model (algorithm) to study the mapping characteristic from the input to the output. In this model, we have a supervisor and a learning agent [Samuel, 1959]. The supervisor can be thought of as a teacher who has the desired output set. When a new input is fed in, the algorithm provides the actual output which is compared with the desired output by the supervisor and is corrected. This is an iterative process which stops when a good level of performance is achieved [Samuel, 1959].

Supervised learning uses classification and regression models. Classification Techniques are used to identify the discrete class each register belongs to by using one or more independent variables. For example, to determine whether an email is spam or not spam, to identify numbers and letters, etc. [Samuel, 1959]. Some algorithms to perform classification are decision trees, neural networks, support vector machines, K-nearest neighbour, etc. Regression Techniques are predictive modelling methods used to predict continuous responses. For example, the temperature in each point, the sales of a given product, time series modelling, etc.

In unsupervised learning, we have only input data and no predefined output data is provided. In this model, we have only the learning agent that is fed with data and here no supervisor is present and thus the computer must learn without guidance. The learning agent itself should find the pattern from the data. Unsupervised learning is more complex, but it can also solve more complex problems [Agrawal, et al., 1993]. The most common unsupervised learning method is clustering. That may be used to find similar groups of information in large information sets. The aim is to find groups in which the objects of each group are comparatively more similar to objects of that group than those of other groups.

In Reinforcement Learning [Sutton and Barto, 2018] the agent learns to behave in an environment by performing actions and seeing what the results of its actions are. The agent gets positive feedback, for good actions and negative feedback or penalties by performing bad actions and learns automatically using feedbacks without any labelled data.

A large number of practical applications emerged for each of these machine learning types. For Classification, applications range from image classification, fraud detection, diagnosis, to customer segmentation. Regression is used for weather forecasting, market and sales forecasting, population growth or life expectancy prediction among many other applications. In the area of Reinforcement Learning several applications with huge recent successes deserve a mention such as game AI, real time decision-making, learning tasks and robot skill acquisition, among many other applications.

## 3.2 Deep Learning

Most of the recent successes of AI have been achieved using deep learning. Deep learning is part of a broader family of machine learning methods based on artificial neural networks. Deep learning architectures use a large number of layers in the neural network, using the multiple layers to progressively extract higher-level features from the raw input. For example, in image processing, lower layers may identify edges, while higher layers may identify the high-level concepts such as letters, digits, or faces [Krizhevsky et al. ,2012] [Bengio et al., 2015].

Deep-learning architectures such as deep neural networks, deep belief networks, deep reinforcement learning, recurrent neural networks and convolutional neural networks have been applied to fields including computer vision, medical image analysis, speech recognition, natural language processing, machine translation, bioinformatics, drug design, climate science, material inspection, board game playing agents and creation of robust robotic skills. In most of these areas deep learning approaches have produced largely surpassing human expert performance [Krizhevsky et al. ,2012] [Bengio et al., 2015].

Deep learning with emphasis for deep reinforcement learning requires a very high degree of computational power. However, the use of deep learning, together with top computational resources enabled some of the most prominent recent successes of AI. These successes include the programs developed by the artificial intelligence research company Google DeepMind - AlphaGo and AlphaGoZero [Silver et al, 2017] that were able to learn from scratch to play the game of Go at a super human level and AlphaZero capable of mastering the games of chess, shogi and go (and any other similar board game) learning is just a few hours to play the game from scratch [Silver et al, 2018].

## 3.3 Deep Reinforcement Learning

Reinforcement Learning (RL) is a feedback-based Machine Learning technique in which an agent learns to behave in an environment by performing actions and seeing what the results of its actions are. For each good action, the agent gets positive feedback, and for each bad action, the agent gets negative feedback or penalty. In Reinforcement Learning, the agent learns automatically using feedbacks without any labelled data. Since there is no labelled data, so the agent is bound to learn by its experience only. RL solves a specific type of problem where decision making is sequential, and the goal is long-term, such as game-playing, robotics, etc.

RL is quite different from supervised learning. In interactive problems it is impractical to obtain examples of desired behaviour and in uncharted territory, an agent must learn from its own experience. RL is quite different from unsupervised learning since it tries to maximize a reward signal, not to find hidden structure in collections of unlabelled data. RL explicitly considers the whole problem of a goal-directed agent interacting with an uncertain environment, creating a behaviour model while applying it in the environment. RL is the closest form of ML to the kind of learning humans do. Typical applications range from auction bidding, to playing games, to robotic automation, going through strategy optimization in multiple areas. The landscape of algorithms in modern RL is displayed in figure 3.

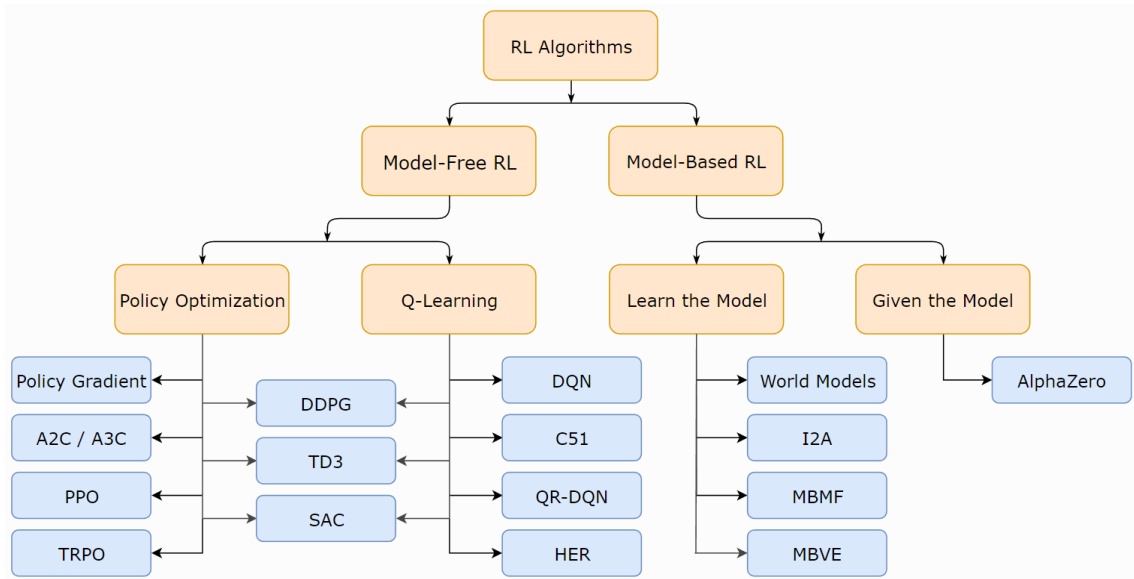

*Figure 3: A simple taxonomy of modern Reinforcement Learning Algorithms (adapted from [OpenAI, 2018])*

With the huge development of new DRL algorithms in the last few years it is very difficult to create a full taxonomy of algorithms in modern RL, because the modularity of algorithms is not well-represented by a tree structure [OpenAI, 2018]. Also, new algorithms including advanced exploration, transfer learning, meta learning and other subjects arise every year.

One of the most important branching points in an RL algorithm is the question of whether the agent has access to (or learns) a model (e.g., a function which predicts state transitions and rewards) of the environment. Having a model of the environment allows the agent to plan by thinking ahead, seeing what may happen depending on its choices, and explicitly deciding between its options and using this for learning the policy such as it happens for example in AlphaZero, resulting in a significant improvement in sample efficiency over model free methods [OpenAI, 2018]. The main problem of model-based RL is that the agent typically has not available a model of the environment and thus to use this approach it has to learn the model by experience, which creates several complex challenges. The biggest challenge is that bias in the model can be exploited by the agent, resulting in an agent which performs well with respect to the learned model, but behaves sub-optimally or very badly in the real environment [OpenAI, 2018].

Model-free methods, although losing in the potential gains that sample efficiency gives from using a model, tend to be easier to implement and tune and have increasingly gain dominance in what concerns DRL. Another critical branching point in the RL algorithm classification is what to learn: stochastic or deterministic policies, action-value functions Q-functions, value functions, and/or environment models [OpenAI, 2018].

In what concerns model-free RL there are now two main approaches for representing and training agents: Policy Optimization with explicitly policy representations and optimization of the parameters by gradient ascent on the performance objective or indirectly, by maximizing local approximations of the performance objective [OpenAI, 2018]. This optimization is almost always performed on-policy and each update only uses data collected while acting according to the most recent version of the policy. Policy optimization also needs learning approximator for the on-policy value function.

A2C and A3C are very good algorithms for policy optimization performing gradient ascent to directly maximize performance [Mnih et al., 2016]. PPO is also excellent on this matter and its updates indirectly maximize performance, by instead maximizing a surrogate objective function which gives a conservative estimate for how much the performance objective will change as a result of the update [Schulman et al., 2017]. Q-Learning and in particular DQN that started the field of DRL learns an approximator for the optimal action-value function and use an objective function based on the Bellman equation [Mnih et al., 2013]. This optimization is almost always performed off-policy, which means that each update can use data collected at any point during training, regardless of how the agent was choosing to explore the environment when the data was obtained [Mnih et al., 2013].

The main RL algorithms developed and used in practice include, among others: DQN - Deep Q-Networks [Mnih et al., 2013], A2C/A3C - Asynchronous Advantage Actor-Critic [Mnih et al., 2016], TRPO - Trust Region Policy Optimization [Schulman et al., 2015], PPO - Proximal Policy Optimization [Schulman et al., 2017], SAC (Soft Actor-Critic) [Haarnoja et al., 2018], HER - Hindsight Experience Replay [Andrychowicz et al., 2017] and AlphaZero [Silver et al, 2018].

As said before, this work focusses on the use of Machine Learning with emphasis for Deep Reinforcement Learning to create advanced, robust, and efficient robotic skills. Although have used almost all the algorithms briefly described at this section most of the more advanced work in the last years, was performed with PPO and several variations of it, including our own variations and implementations of this algorithm.

# 4. Intelligent Robotics

## 4.1 Robots and Robotics

A robot may be defined as a programmable machine capable of carrying out a complex series of actions automatically. In fact, the term "Robot" comes from a Slavic with meanings associated with slavery labour. The word "Robot" was first used to denote a fictional humanoid in a 1920 Czech-language play by Karel Čapek. Robots can be autonomous or semi-autonomous and range from humanoids such as Honda's ASIMO to industrial robots, medical robots, assistive robots, swarm robots, autonomous vehicles, flying drones and even microscopic nano robots.

Robotics is the branch of science and technology that deals with the project, construction, programming and application of Robots. It is basically the study of Robotic Agents (e.g. agents with body). This poses an increased complexity to the development of AI compared with traditional applications. The robotics environment is the real world that is a lot more complex than traditional environments for AI application. The real world is dynamic, only partially accessible, continuous, non-deterministic and multi-agent. The robot must gather its own perception by means of a series of sensors and interpret it using computer vision, sensor fusing techniques among many others. The robot action may be also quite complex. For example, for typical humanoid robots, more than 20 motors must be controlled in real-time for the robot to be able to stand-up and walk. Researchers must also deal with the Robot physical and control architectures, with the need to navigate in typically unknown environments, map the environment and localize itself on it, and the need to interact with other robots or humans and even cooperate with them to perform complex tasks. A number of excellent text books exist describing the area of robots and robotics including among many others [Thrun el al., 2005], [Choset et al., 2005], [Siegwart et al., 2011], and [Murphy, 2019].

## 4.2 Robotics Applications

Robots were mainly used to perform dangerous or tasks that could be very difficult to be performed directly by humans and/or repetitive tasks that may be performed a lot more efficiently (or cheap) by robots than when performed by humans. However, robotics has evolved a lot in the last few years and robots have moved from manufacturing, industrial applications to domestic Robots (pets like the Furby or the AIBO, vacuum cleaners), entertainment and social robots, medical and personal service robots, military and surveillance robots, educational robots, autonomous vehicles (autonomous cars, ships, submarines, airplanes, drones), new industrial applications (such as mining, fishing and precision agriculture), exoskeletons and active orthoses, ambient assisted living, new hazardous applications (like space exploration, military apps, toxic clean-up, mine cleaning, construction and underwater applications) and more recently multi-robot applications and the start of the creation of Human-Robot Teams. Many of today's robots are inspired by nature contributing to the field of bio-inspired robotics. The use of robots in military combat and other critical applications raises ethical concerns. The possibilities of robot autonomy and potential repercussions have been addressed in fiction and are already a realistic concern.

## 4.3 Simulation for Robotics

In order to be able to use machine learning for robotics simulation is essential. Simulation may be defined as "Imitation of some real thing, state of affairs, or process, over time, representing certain key characteristics or behaviours of the physical or abstract system". Its applications are huge, such as: Understanding the system functioning; Performance optimization; Testing and validation; Decision-making; Training and education; Testing future/expensive systems; and mostly to accelerate time! Simulation should be used mostly for complex systems impossible to solve mathematically. For using machine learning in robotics simulation is needed. In fact, simulation enables to compress/accelerate time which, associated with the availability of very good computational resources is decisive for performing the, typically needed, millions of iterations for using reinforcement learning algorithms for practical applications. However, the simulators must be realistic enough so that the skills developed in simulation may be ported to real robots and bridging the gap between simulation and robotics is one of the main research open issues on this area.

## 4.4 Multi-Robot Coordination

Recently, Multi-Robot Systems (MRS) have attained considerable recognition because of their efficiency and applicability in different types of real-life applications. A series of studies have been made concerning MRS and MRS coordination, trying to define basic terminology, categorization, application domains, and coordination approaches for each application domain [Verma and Ranga, 2021].

There are numerous historical survey papers concerning Multi-Robot Systems and Multi-Robot Coordination such as [Matarić, 1995], [Dudek et al., 1996], [Stone and Veloso, 2000], [Iocchi, et al., 2001], [Gerkey and Matarić, 2004], [Farinelli et al., 2004] or [Ota, 2006]. Although several survey papers have been published in the past related to MRS, however, only a few are related to MRS coordination, although there is large number of works being performed in this area. Recent surveys on MRS coordination and cooperation include [Ismail and Sariff, 2019], [Rizk et al., 2019] and [Verma and Ranga, 2021]. In this latter survey MRS are classified concerning the level of coordination, the composition of the team including its heterogeneity, the type of environment and its cooperative/competitive characteristics, the type of communication available, the reactive or deliberative architectures, and the team size.

Coordination in MRS may be defined as working in harmony in a group of robots. Coordination may be used as the mechanism for cooperation although a set of robots may work together without any type of cooperation. Given a task, a multi-robot system displays cooperative behaviour if, due to some coordination mechanism there is an increase in the total utility of the system [Cao et al., 1997]. In cooperation, not only robots pay attention to their own actions and goals but also need to know if there are other task with priority from the teammates and thus the actions performed by each robotic agent must consider the actions executed by the other robotic agents to select the actions to maximize the global system efficiency or utility. Coordination may be classified based on various parameters, such as communication mode, decision making, adaptivity, and protocol [Verma and Ranga, 2021]. A large number of issues and challenges is available concerning coordination of MRS such as the Communication Model, Explicit and Implicit Communication, Heterogeneity, Scalability, Robustness and Energy Efficiency, Internet of Robotic Things, Multi-Robot Learning, Human-Robot Interaction, among many others.

## 4.5 Robot Learning and Multi-Robot Learning

The Robot Learning and Multi-Robot Learning fields have multiple open problems and challenges, some of which inherited from single-agent learning. These include the trade-off between exploration and exploitation, the integration of domain knowledge, problem decomposition, credit assignment and the curse of dimensionality. The MAS perspective also creates challenges, such as an adequate learning goal, non-stationary environments, communication, and coordination. The exploration and exploitation trade-off, where either the algorithm tries out new actions to measure their effectiveness or exploits actions that are already known to yield a high reward. If the focus on either one is too strong, learning will yield poor results, and this is especially critical when dealing with real or simulated robots. Integration of domain knowledge is usually in the form of problem modelling or adequate initial solutions, which have been shown to increase the learning speed of agents. Credit assignment, from a single-agent perspective, is based on how to properly reward agents for their actions when rewards are not immediate, and from a multi-agent perspective, is based on how to properly reward agents who did not contribute equally to the task completion. The curse of dimensionality relates to the exponential growth in complexity seen in partially observable Markov decision process and in MAS. Adequate learning goals and non-stationary environments related to the adaptation and stability properties of Multi-Agent Reinforcement Learning (MARL) algorithms. Communication and coordination arise as the main challenges to solve in the field of MAR. It is of great importance to define functional communication protocols for the learning phase that enable to improve the learning process speed. However, care is needed since after deploying the solution. in the real world, this communication, most probably, will no longer be available and thus the learned solution should not be dependent on it.

## 5. RoboCup and Robotic Soccer

### 5.1 RoboCup History

The year of 1997 is remembered in the areas of Artificial Intelligence and Robotics as a decisive point in their history. In May 1997, Deep Blue from IBM defeated Gary Kasparov, the human world champion in chess successfully ending a long challenge for the AI community. On July 1997, NASA's MARS Pathfinder made a successful landing, and the first autonomous robotics system was deployed on the surface of Mars. Together with these accomplishments, the first RoboCup competition, RoboCup1997 was held enabling worldwide researchers in AI and Robotics to participate, for the first time, in a robotic soccer competition (real and simulation) taking the first steps toward the development of robotic soccer players which can beat, in the future, the human World Cup champion team [RoboCup, 1997], [Kitano et al., 1997].

The idea of robots playing soccer was first mentioned by Alan Mackworth (University of British Columbia, Canada) in 1992, later published [Mackworth, 1993] and subject to several publications in the scope of the Dynamo project at Canada. Independently, a group of researchers at Japan organized a Workshop on Grand Challenges in AI in October 1992 in Tokyo, including a discussion of using soccer for promoting science and technology and AI research. In June 1993, Minoru Asada, Hiroaki Kitano and other Japanese researchers, after conducting a series of studies, decided to launch a robot soccer competition named Robot J-League but that, following a large number of international requests was extended as an international joint project and renamed as RoboCup - Robot World Cup Initiative [RoboCup, 1997].

Itsuki Noda, at the ETL Laboratory in Japan, was also conducting multi-agent research using soccer, at the time, and started the development of a multi-agent soccer simulator that later became the official soccer simulator of RoboCup. The server was then updated and extended by a large RoboCup community and is still used today for the RoboCup soccer simulation 2D competition. Also, Minoru Asada's Lab at Osaka University in Japan and Manuela Veloso together with Peter Stone at Carnegie Mellon University in the USA were also already working on soccer playing robots. In September 1993, the first announcement of RoboCup was made, together with initial competition rules. Several discussions on organization and technical issues were held at numerous conferences and workshops. Meanwhile, Noda's team at ETL announced the Soccer Server as an open multi-agent soccer simulator freely available.

During IJCAI'95 held at Montreal, Canada, the First Robot World Cup Soccer Games and Conferences event was announced to be held together with IJCAI'97 in Nagoya [RoboCup, 1997]. Also, a first demonstration of Noda's simulator was performed. A pre-RoboCup-96 event was also held during IROS'96, Osaka, in 1996, with eight teams competing in a simulation league and robotic demo of the middle size league. The first official RoboCup games and conference was held at Nagoya in 1997 with great success. Over 40 teams participated (real and simulation combined), and over 5,000 spectators attended [RoboCup, 1997].

Since then, RoboCup grew significantly and there are more than 1000 major teams and 10000 junior teams active at the moment [Asada and von Stryk, 2020]. RoboCup features also a large number of local/regional events that qualify teams for the main competition such as The RoboCup Asia Pacific, RoboCup German Open, RoboCup Portuguese Open, RoboCup Chinese Open, RoboCup Iran Open and RoboCup Latin America/Brazil Open. In Europe the RoboCup German and Portuguese Opens are used as the European competitions qualifying teams for the main competitions in different leagues distributing the main competitions between the two opens [RoboCup, 2022].

RoboCup had already 25 editions and is one of the largest world's Robotics/AI events in the world with typically more than 3000 participants. RoboCup 2020 was not held due to the COVID-19 pandemics. However, RoboCup 2021 took place as a successful distributed remote event on June 22 - 28, 2021, featuring more than 2000 participants [Stone, 2021]. RoboCup 2021 online was hosted by Underline (https://underline.io/events/108/reception), which provided both the virtual environment for RoboCup Symposium, opening ceremony, award ceremony, and sponsor booths and also Gather Town (https://www.gather.town/) virtual space for social interaction [Stone, 2021]. In 2022 it was possible to restart presential RoboCup competitions and RoboCup was held at Bangkok, Thailand, July 2022, with more than 3000 participants [RoboCup, 2022].

## 5.2 RoboCup Main Leagues

The focus of the RoboCup competitions is in autonomous multi-robot soccer playing. Here, the research goals concern in creating cooperative multi-robot and multi-agent systems in dynamic adversarial environments but also single robot perception, action, decision, and skill acquisition. Coordination methodologies, together with robust perception and decision capabilities are essential for creating a successful robot-soccer team. However, in most of the leagues, it is decisive to create robust and flexible individual skills and for the machine learning, with emphasis for deep reinforcement learning is of paramount importance. Besides being very challenging problems for the research in AI, Robotics and ML, the RoboCup league are created also to serve as an opportunity to educate and entertain the general public around science and technology issues concerning AI and robotics.

RoboCup also features four other domains, each comprising several leagues: RoboCupRescue, RoboCupJunior, RoboCup@Home, and RoboCupIndustrial. The RoboCup Rescue competition is aimed at increasing awareness of the challenges involved in search and rescue applications and develop simulators and intelligent robots for rescue scenarios. RoboCup Junior is an educational initiative for young students designed to introduce RoboCup to primary and secondary school children. It offers to the participants the chance to take part in international exchange programs and to share the experience of meeting peers from abroad. RoboCup Junior offers several challenges such as simplified soccer and rescue competitions and an on-stage competition. The RoboCup@Home league aims to develop service and assistive robot technology with high relevance for future personal domestic applications. A set of benchmark tests are used each year to evaluate the robots' abilities and performance in a realistic home environment setting. Focus lies on Human-Robot-Interaction, Navigation and Mapping in dynamic environments, Computer Vision and Object Recognition and Manipulation. Finally, RoboCup@Work is the newest league in RoboCup, targeting the use of robots in work-related scenarios to tackle open research challenges in industrial and service robotics [RoboCup, 2022].

Table 1 (adapted from [Asada and von Stryk, 2020]) displays the RoboCup leagues held since 1997 until 2022 and the author participations and awards. In the table, the Major awards displayed, correspond to winning a trophy ($1^{st}$ – $3^{rd}$ place) or winning the first place at the free/scientific league challenge. These challenges only started in the last 10 or twelve years for most of the leagues.

*Table 1: RoboCup Leagues, Participations and Awards (1997-2022)*

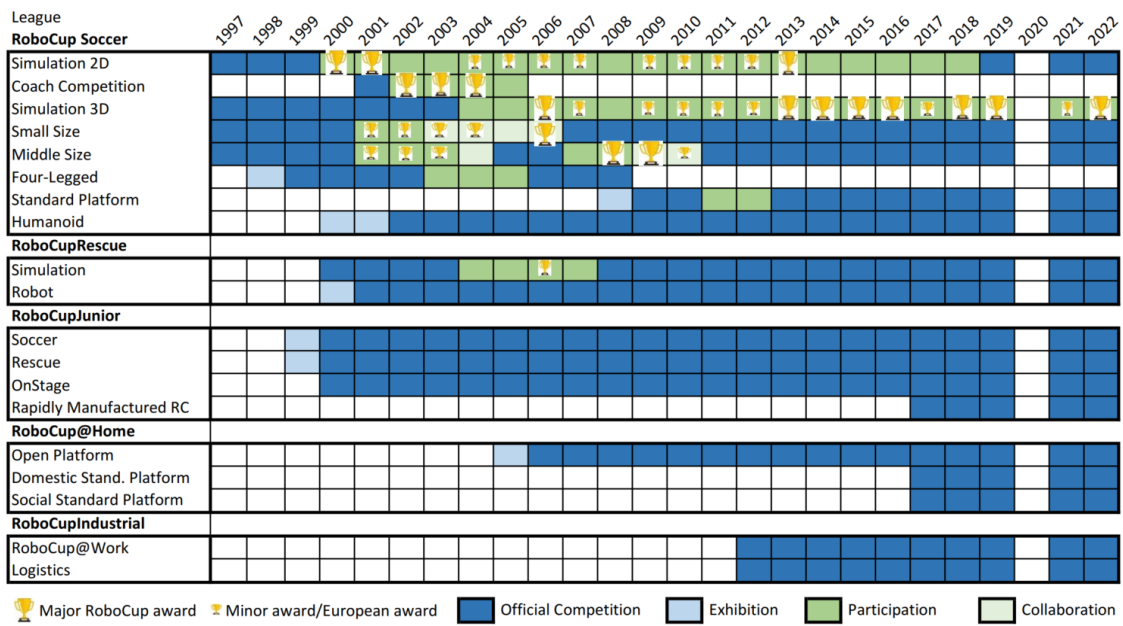

The author research on RoboCup has been focused mostly on the RoboCup Soccer leagues although with some participations also in the RoboCup Rescue simulation league. In the Robotic Soccer leagues the research was applied mostly in the Simulation 2D, Simulation3D, Coach Competition, Small-Size, Middle-Size, Four-Legged League, and Standard Platform league, with successful participation in all these leagues. In the last few years, the research has been focused mostly on robot learning and multi-robot learning for humanoid soccer teams and applied in the soccer simulation 3D league

## 5.3 Humanoid Simulation 3D League

The RoboCup Humanoid Soccer Simulation 3D competition started in 2004 to increase the realism of the simulated environment used in 2D simulation league by adding an extra dimension and more complex physics using a new simulator developed for this purpose – SimSpark with a RCSS3D plugin for simulating the robotic soccer environment [Boedecker and Asada, 2008]. The package is open source and since then is developed by the RoboCup community. From 2004 to 2006, the only available robot model was a spherical agent. In 2007, a simple model of the Fujitsu HOAP-2 robot was made available, being the first time that humanoid models were used in the simulation league. This fact, initially changed the aim of the 3D simulation competition from the design of coordination and strategic behaviours for playing soccer towards the low-level control of humanoid robots and the creation of robot skills like walking, kicking, turning, falling down and getting up, among others [RoboCup, 2022].

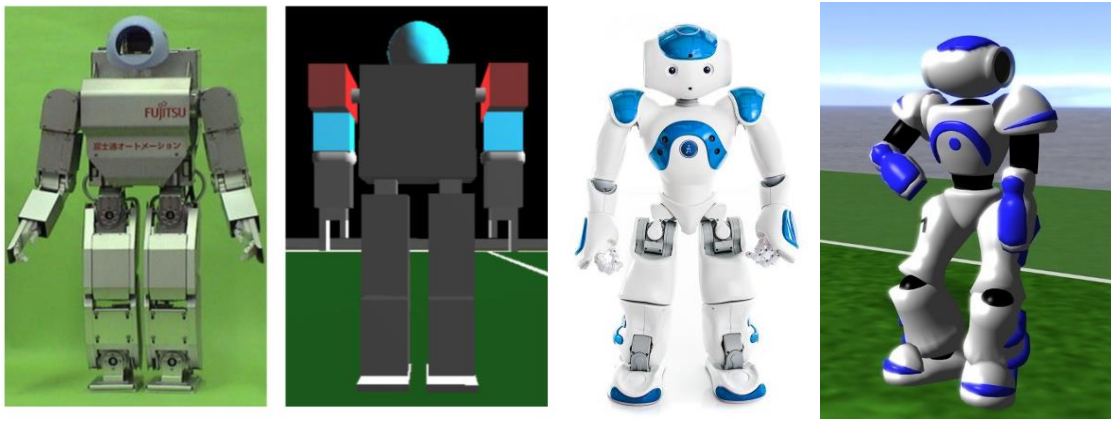

*Figure 4: Fujitsu HOAP-2 Real and Simulated Robot (left); NAO Real and Simulated Robot (right)*

In 2008, the introduction of a NAO robot model to the simulation significantly improved the 3D Simulation League (Figure 4). The real NAO robot, at the time, from Aldebaran robotics was the official robot for the Standard Platform League since 2008. By using a very approximated model for the real and simulation competitions a great opportunity was given for researchers to develop their machine learning and coordination algorithms in the simulation before deploying them into the real robots.

After some initial years focusing on low-level control of the simulated robots, the interest in the 3D simulation competition grew fast and research got back to machine learning and coordination issues such as multi-agent higher-level behaviours based on solid low-level skills for the humanoid robots.

In consecutive years, the number of robots was increased continuously and after starting with 2 vs 2, reached 11 vs 11 in 2012. In 2013, heterogeneous robot types of the standard Nao robot, with different motors and robot part sizes, were introduced. Also, the first drop-in player challenge showed the performance of the teams when playing with unknown teammates of other teams.

In the following years, the league has showed a huge development of the low-level behaviours such as walking, omni directional walking, kicking, passing, getting up and falling down, different goal keeper behaviours, and even running and sprinting and this year very powerful dribbling skills learned using DRL methodologies were shown.

## 5.4 The Challenges of RoboCup for AI

RoboCup has considerably stimulated research carried out in Distributed Artificial Intelligence and Intelligent Robotics. But more than that, it drew the attention of the media and the public to this research. The RoboCup leagues pose distinct but at the same time interrelated research problems [Reis, 2003]. The simulation 3D league includes all the complexities found in real robotic systems such a realistic humanoid robot, realistic sensors and actuators, realistic errors in perception and action. This simulator however, at the same time, uses a very realistic scenario from the point of view of the cooperative task to be performed by the teams of simulated robots - playing a soccer game.

The simulation 3D league where most the work presented in the following sections was applied, poses a wide range of challenges to researchers in the field of Artificial Intelligence, Multi-Robot Systems and Machine Learning. The most important domain characteristics and associated challenges pretended by the simulator include:

- Real-time simulation. The simulator updates the world in discrete but very small time intervals (simulation cycles), each with a duration of 0.02 seconds. The agents receive different types of sensory information and may send requests to execute actions to the simulator each 3 time steps. This requires agents to be able to process all the information received autonomously and reason and act in the environment in real-time.

- Several sources of sensorial information and availability of configurable sensors. Agents have three types of sensors: visual, auditory, and physical. The visual sensor provides agents with information (distances, directions, etc.) about objects in their field of vision (the ball, field lines, goal post, robot visible parts and other robot parts, etc.). Visual information is incomplete (agents have reduced vision cones), it is provided in an agent-relative perspective and with significant errors. The auditory sensor provides agents with information regarding messages issued by other agents, but with limited range and very small message size. The physical sensor provides information about the player's joints, contact with the ground, and the robot acceleration and velocity.

- Unreliable and low-bandwidth communication. Contrary to what happens in most MAS applications, in this domain, agents cannot communicate directly with each other. Communications are carried out through the simulator using the say and hear protocols that restrict communication in various ways. In addition to this fact, the 22 agents use the same communication channel, which has low bandwidth and is very unreliable. Agents only hear a single message per cycle and thus they need to think about the interest to speak. Also, agents may prioritize hearing messages from a given teammate enabling the creation of more intelligent team communication protocols.

- Multi-objective, partially cooperative and partially adverse environment. Agents have two contradictory high-level objectives: defending to avoid conceding goals and attacking to score goals. The environment is partially cooperative, and agents must behave as members of a team. It is also partially adverse, in the sense that there is another team of agents with exactly the opposite objectives.

- Need to transform very low-level actions (use the motors to move the robot joints) into high-level skills such as walking, kicking, or getting up from the ground. In order to perform a certain simple action, such as kicking the ball in a given direction, dribbling the ball or intercepting a moving ball, it is necessary to perform a sequence of very large number of low-level elementary actions.

- Impossibility of generating competitive skills by hand. Giving the complexity of the environment and the 23 degrees of freedom of the simulated NAO robot, it is completely impossible to generate competitive skills by hand. Thus, this makes this domain as an excellent playground for machine learning with emphasis on deep reinforcement learning algorithms.

- Need to create complex collective actions. The complexity of the cooperative task to be performed (play a 11 vs 11 game of soccer) implies the need to create, in real-time, complex collective actions such as passes, individual marking, collective plays (involving the exchange of the ball between several players), collective defensive movements (involving two or more players), definition of formations enabling the team to move, correctly occupying the field, etc.

- Automated Referee awarding fouls for crowding and pushing. In 2011 we created a new automated referee agent capable of referring the game assigning fouls for crowding and pushing and removing the offending players from the field. The players removed may then renter by the opposite side line of the field. The referee was then extended by the community and since then suffered several improvements and is the official referee of the simulation 3D games. The automated referee poses several challenges to the robot movement such as the need to avoid pushing the opponents and to be able to navigate avoiding obstacles.

- Heterogeneous Robots. The introduction of heterogeneous robots in the league brought additional challenges to the domain. Each team can choose its players from 5 different robot types with slightly different bodies and capabilities. The team must use at least three different robot types which makes it even more complex to use hand made skills and enforces the need for flexible algorithms optimised with machine learning.

The next chapter will show how we deal with the complexities and challenges of the chosen environment and the coordination and machine learning algorithms we developed for tackling this problem.

# 6. Coordination and Machine Learning in Robotic Soccer

This section briefly describes some of the coordination and machine learning methodologies developed by the author and his team, with emphasis on the Robotic Soccer team FCPortugal, in projects coordinated by the author and that were successfully applied to RoboCup. Most of the works developed are only described at a very high level but pointers to the specific papers where details may be found, are included.

## 6.1 Strategy, Tactics, Formations and SetPlays

This work was aimed at creating the concept of a strategy for a (robotic) soccer teams based on tactics, formations, setplays, player types, active and passive behaviours, and concepts such as Situation Based Strategic Positioning and Dynamic Positioning and Role Exchange, among several other coordination strategies for robotic soccer teams. The content of the section is based on the work that led, among others, to the following main publications: [Reis et al., 2001a], [Reis et al., 2001b], [Lau, Reis, 2001], [Reis, 2003], [Lau et al., 2007], [Mota et al., 2008], [Almeida et al., 2010], [Reis, 2012], [Reis, 2013], [Reis et al., 2013], [Reis, 2017].

CMUnited brought the concepts of formation and positioning to RoboSoccer [Stone, 1998] [Stone and Veloso 1999] [Stone et al. 2000] and used a very simple dynamic switching of formations as well. FCPortugal extended these concepts and introduced the concepts of tactics and player types. FC Portugal's team strategy (Figure 5) is based on a set of tactics to be used in different game situations and a set of player types. Tactics include several formations used for different game specific situations (defence, attack, goalie free kick, scoring opportunity, etc). Formations are composed by eleven positionings that assign each player a given player type and a base strategic position on the field. One of the most significant features is the clear distinction between strategic situations (when the agent believes that it is not going to use an active behaviour soon) and active situations (ball recovery and ball possession). In strategic situations, players use a Situation Based Strategic Positioning. For active situations - ball possession, ball recovery or game stopped - decision mechanisms based on the integration of real soccer knowledge are used.

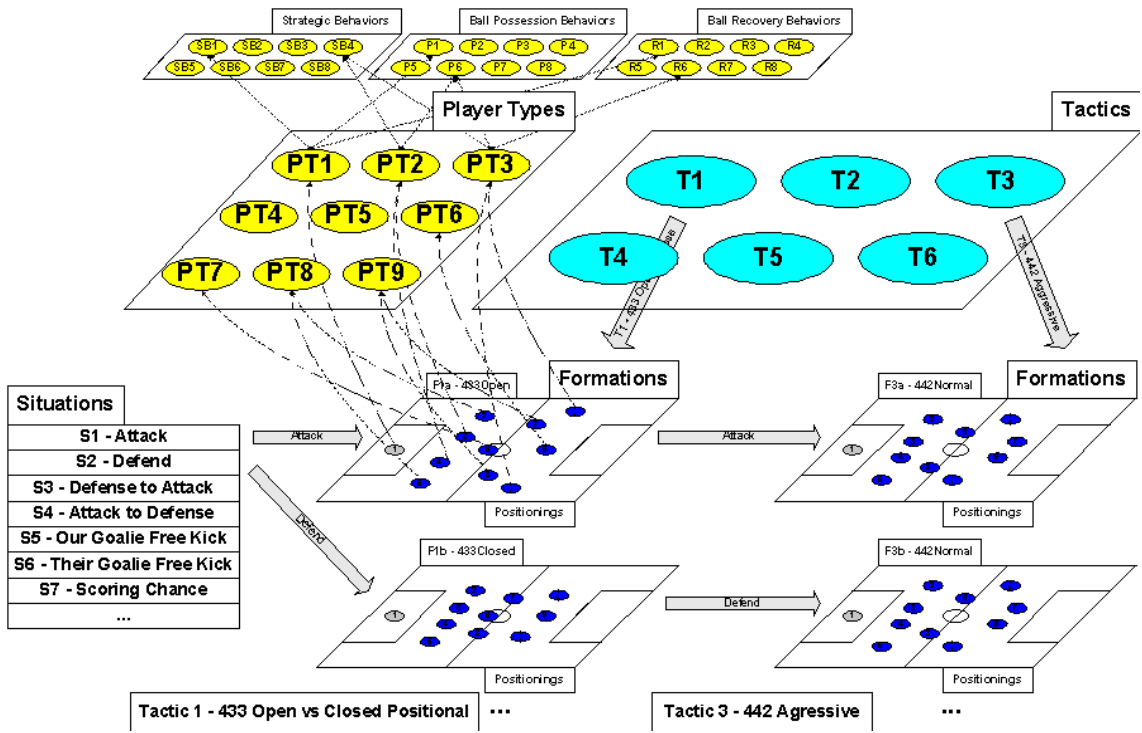

*Figure 5: Strategy Definition for a Robotic Soccer Team*

Situation Based Strategic Positioning (SBSP) mechanism [Reis and Lau, 2001] [Reis et al. 2001] is used for strategic situations (in which the agent believes that it is not going to enter in active behaviour soon). To calculate its strategic positioning, the agent analyses which is the game situation. Then the agent calculates its base strategic position in the field in that formation, adjusting it according to the ball position and velocity, situation, and player type strategic information. The result is the best strategic position in the field for each player in each situation. Since, at each time, only a few players are in active behaviour (conducting the ball or trying to recover the ball) most players are close to their strategic positionings. SBSP enables the team to move similarly to a real soccer team, covering the ball while the team remains distributed along the field. Formations were latter extended by [Akiyama and Noda, 2007] with a formalization of a formation as a map from a focal point like a ball position in a soccer field to a desirable positioning of each player agent, and a proposal to approximate the map using Delaunay Triangulation. Based on this work we further extended our SBSP mechanism with a graphical application enabling to define formations for each situation using Delaunay Triangulation and using not only the ball position but also local information to further optimize the strategic position.

The Dynamic Positioning and Role Exchange (DPRE) [Reis and Lau, 2001] [Reis et al. 2001], was based on previous work from Peter Stone [Stone, 2000] which suggested the use of flexible agent roles with protocols for switching among them. The concept was extended, and players may exchange their positionings and player types in the current formation if the utility of that exchange is positive for the team. Positioning exchange utilities are calculated using the distances from the player's present positions to their strategic positions and the importance of their positionings in the formation on that situation. This concept was later extended in our simulation 3D league. Figure 6 shows our Matchflow application for the graphical definition of formations in any type or robo(soccer) league.

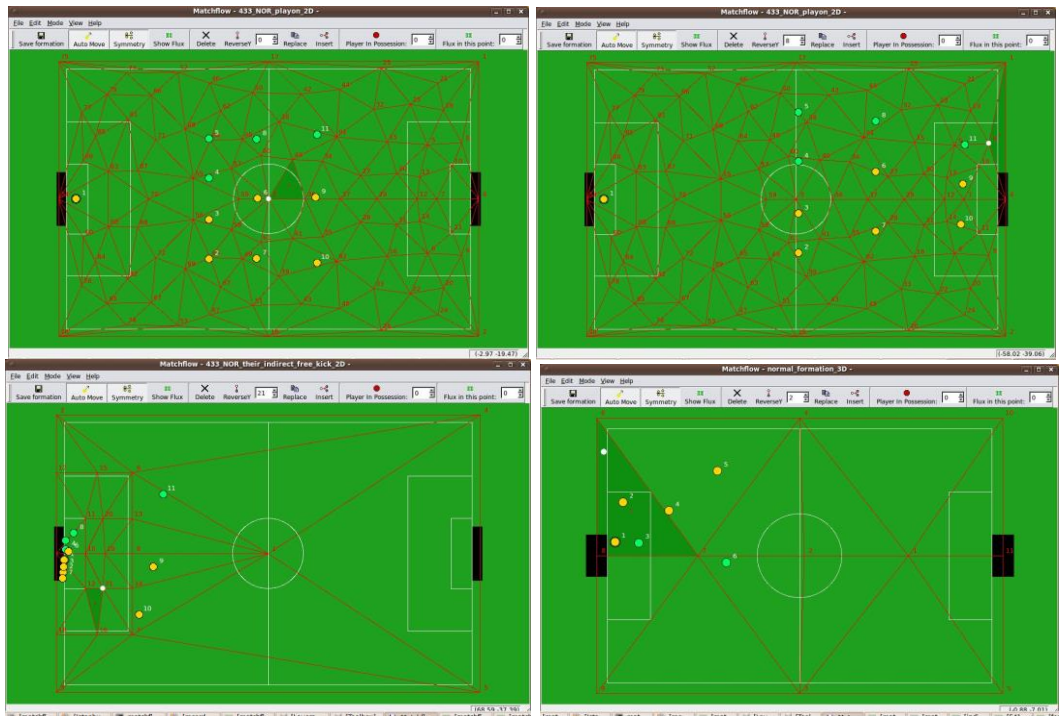

*Figure 6: Matchflow- Graphical Definition of Formations*

One of the main developments of FC Portugal team is the improvement of its ability to be coached both before and during the game. Before the game, the strategy is defined through configuration files and during the game the tactics used may be changed by the automatic coach or even without the coach intervention by a set of rules known by all players. The definition of strategic behaviours has been enhanced with new parameters that effectively control player movements. Active behaviours have also been extended to support different playing styles. Some of the concepts of Coach Unilang like game pace, team aggressiveness, team mentality, etc have been included in order to turn the team adaptable to the opponent's strategy.

For each game a human coach specifies the strategy to be taken by the team. This strategy consists of several tactics that should be activated according to the opponent team' behaviour and statistical information gathered during the game. Each tactic consists of several global parameters that control the team behaviour and of several formations that are active in different situations (like defence, attack, etc). For each formation the human coach can specify the strategic and active behaviour of each player. During the game players follow the coach advice to change tactics, formations and player types. Players may also decide tactical changes by themselves, based on statistical information sent by the coach and teammate and opponent modelling techniques.

Throughout the years the flexible tactical approach has been extended and now, the team configuration files are far more flexible, enabling to change completely the team behaviour for a given game. Several concepts were added such as setplays and now, each tactic contains a series of setplays that are used for different situations. The strategy also includes the concepts of flux, safety and easiness. The flux is concerned with the value of each point on the field. It is defined in the same graphical application as the formations and also uses Delaunay Triangulation for defining the flux for all possible point on the field. Each action changes the value of the flux from the initial point to the target point. Each action has a given safety that is concerned with the possible scenarios of the success of the action. For example, an action to pass the ball to the own goalie is very unsafe because if unsuccessful may end on an own goal. Finally, each action has a given easiness. For example, having to round the ball to kick it in the opposite direction is a lot more difficult than kicking the ball to the front. The strategy combines these three concepts, and each tactic has a given weight for flux, safety, and easiness. If the weight for the flux is 1 and the other weights 0, the players will only try to shoot the ball to the opponent goal. If the safety is one, the team will only keep the ball. This enables great flexibility for the team behaviour.

## 6.2 Team Coaching and Coaching Languages

Our work on Coaching was aimed at creating the concept of a strategy for a (robotic) soccer game and a language to communicate with the robotic soccer team managing its behaviour at a very high-level. The content of the section is based on the work that led, among others, to the following main publications: Coach [Reis, Lau, 2002a] [Reis, Lau, 2002b] [Reis, Lau, 2002c] [Sampaio et al., 2004]

The FC Portugal team developed COACH UNILANG, a standard language for coaching (Robo)Soccer teams. This language was developed with the main objective of coaching FCPortugal teams. The language enables high-level and low-level coaching through coach instructions. High-level coaching includes changing tactics, formations used in each situation and changing player behaviour. Low-level coaching includes defining formations, situations, player behaviour and positioning with high detail. The language also enables the coach (functioning like an assistant coach) to send opponent modeming information and game statistical information to the players. The language was the base of the RoboCup Coach Competition with most of its features being imported to the official RoboCup Coaching Language between 2021 and 2004.

COACH UNILANG was the first attempt to create a language that enables high-level coaching of robo(soccer) teams. Another attempt to create a standard language for coaching robosoccer teams was made by Patrick Riley, Gal Kaminka and Timo Steffens but this later language did not include high-level concepts like tactics, formations, player types, time periods or situations (essential to define the behaviour of a soccer team). It only enabled the definition of player's home positions and several directives concerning individual action behaviours (like which player to pass the ball or which player to mark). Also, this language included some concepts derived from COACH UNILANG (whose draft proposal was made public in January 2001) like definitions for conditions, actions and directives, marking passing lanes and playmode based conditions. This "standard language" was used in a "coaching competition" in Seattle. As expected, no visible tactical changes were seen in any of the teams that did use this "standard language" and the competition was considered by most of the participants as "somewhat random". Opposite, FC Portugal 2001 did change its tactic and even played, when losing near the end of a game, in a very aggressive 443 formation (without a goalie)! Some people may argue that the definition of tactics and formations is too high-level. However, we argued that COACH UNILANG enabled the definition of tactics and formations at any level. The coach may use the high-level instruction mechanisms, or he may define a tactic, defining all used formations and the behaviour of each player used in each formation at low level (defining for each region what are the player preferred actions and configuring them). This enables virtually the definition of any soccer tactic. Other critique that can be made to this language is that it is too complex. However, soccer is a complex game. So, there is no interest in developing a language that is able to train only teams that use simple decision mechanisms and is not able to play like real soccer teams (capturing and using the real soccer complexity). FC Portugal, even with a competition that used a simple language without the high-level features needed to coach a (robo) soccer team, won the coach competition in 2002 and got second place in 2003 and 2004.

## 6.3 SetPlays – Multi-Agent Coordination and Strategic Planning

Our work on Setplays was aimed at creating the concept of a Setplay for a (robotic) soccer game and a graphical application to define flexible setplays with application in soccer in general and in any robo(soccer) league. The content of the section is based on the work that led, among others, to the following main publications: [Mota et al., 2006], [Mota et al., 2007a], [Mota et al., 2007b], [Reis et al., 2010], [Mota et al., 2010a], [Mota et al., 2010b], [Mota et al., 2011], [Mota, 2012], [Fabro et al., 2014], [Mota et al., 2014], [Cravo et al., 2014].

Multi-agent coordination and strategic planning are two of the major research topics in the context of RoboCup. However, innovations in these areas are often developed and applied to only one domain and a single RoboCup league, without proper generalization.

A Setplay is a freely definable, flexible and multi-step plan, which allows alternative execution paths, involving a variable number of robots and having a set of activation, finish and abort conditions. Also, although the importance of the concept of Setplay (or Set Piece), to structure the team's behaviour, has been recognized by many researchers, in the area of sports but also in the area of multi-agent systems, no general framework for the development and execution of generic Setplays had been presented in the context of RoboCup. Figure 7 shows our Setplays High-Level Definition Diagram.

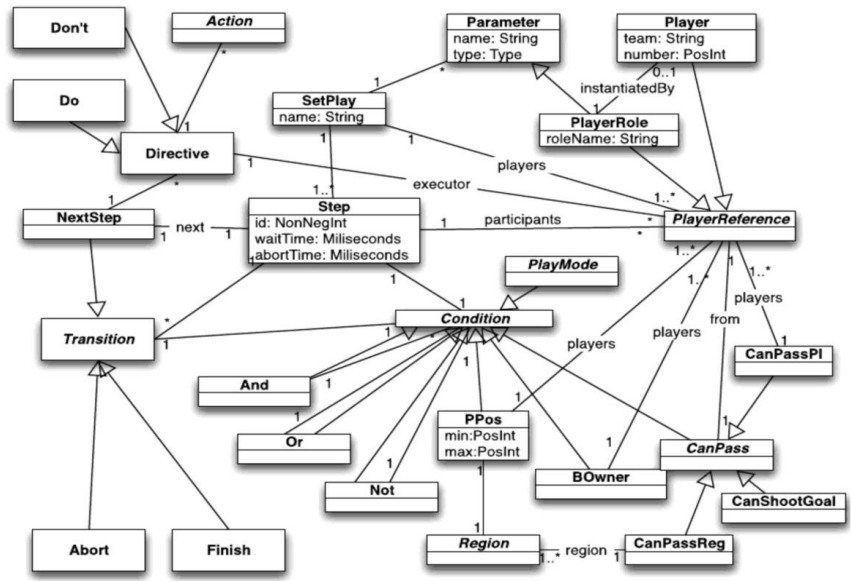

*Figure 7: Setplays High-Level Definition Diagram*

We developed a framework for high-level Setplay definition and execution, applicable to any RoboCup cooperative league and similar domains. The framework is based on a standard, league-independent and flexible language, based on S-Expressions that defines Setplays, which may be interpreted and executed at run-time through the use of inter-robot communication. A real-time selection algorithm for Setplays, inspired by Case-based Reasoning (CBR) techniques, was also developed. This tool stores the past application conditions and success of individual Setplays and uses this knowledge to select among Setplays whenever they are feasible. The Setplay Framework aims at being used as a tool to rapidly prototype multi-agent plans, that are executed at run-time, allowing the swift adaptation to particular opponents, and thus exploiting weaknesses and gaining a competitive edge.

The work developed also showed us the need to study methods of representation and visualization of set plays that allows their definition in a consistent way with the framework previously developed and also the development of a graphical application for designing these same representations. Thus, we have developed a graphical application (Figure 8) for graphically defining setplays and that provides interconnection from the Setplay framework with other software for testing and debugging setplays. The application's usability tests, gave very positive results, especially in the reduction of both wasted time and number of errors committed in the definition of set plays compared with the manual definition. At the end of the project, it was concluded that the methodologies used to represent and display set plays were well chosen, resulting in a very effective application for design, testing, debugging and improving set plays. The developed application was used since its development as the best solution for the definition of set plays for the team FCPortugal and was subject to several minor developments afterwards.

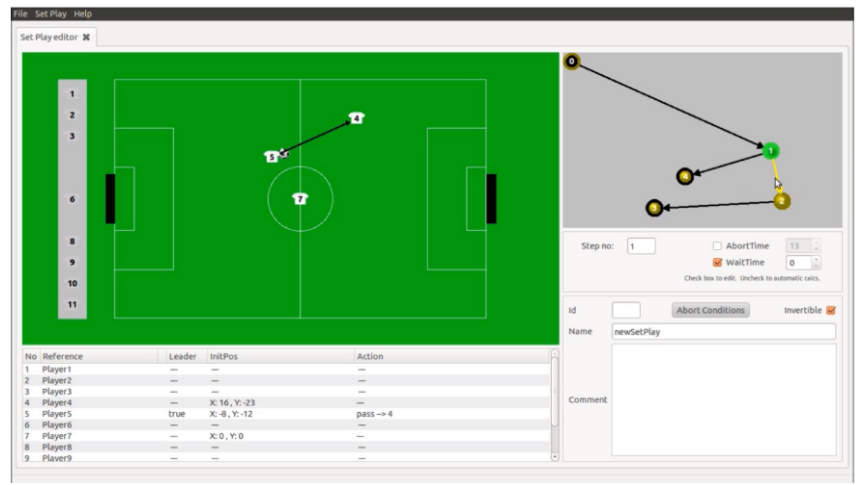

*Figure 8: Strategy Planner - Graphical Definition, Test and Adjustment of Setplays*

The framework and graphical application were applied in the 3D and 2D Simulation Leagues, as well as in the Middle-Size League. Concrete Setplays were created and included in the strategy of the teams in the three abovementioned leagues, allowing the framework to be properly evaluated. The results achieved show the usefulness of this approach. Experiments were made in competitive settings, in the scope of the 2D, 3D and MSL leagues. The results showed a clear improvement in the performance when Setplays are activated. This motivates the usage of the Setplay Framework as a main coordination technique of any team participating in the Simulation, Small-Size, Middle-Size, Humanoid and Standard Platform leagues of RoboCup. Our work on Setplays also included in the last few years the optimization of Setplays using Machine Learning.

We have also developed an approach based on Reinforcement Learning (RL), that allows the use of experience to devise the better course of action in set plays with multiple choices. Simulations results showed that the proposed approach allows a team of simulated agents to improve its performance against a known adversary team, achieving better results than previously proposed approaches.

## 6.4 Strategy Optimization using Machine Learning

This work was aimed at optimizing high-level feature of the FC Portugal code, including, tactics, formations and setplays using machine learning. The content of the section is based on the work that led, among others, to the following publications: [Gonzalez et al., 2008] [Almeida et al., 2009a] [Almeida et al., 2009b] [Abreu et al., 2010] [Faria et al., 2010a] [Portela et al., 2010] [Faria et al., 2010b] [Abreu et al., 2010] [Faria et al., 2010c] [Almeida et al., 2010] [Faria et al., 2011] [Abreu, 2011] [Almeida et al., 2012] [Reis, 2012] [Abreu et al., 2012a] [Abreu et al., 2012b] [Almeida et al., 2013] [Abreu et al., 2013] [Reis, 2013] [Reis et al., 2013] [Fabro et al., 2014] [Abreu et al., 2014] [Reis, 2017] [Abreu et al., 2019a] [Abreu et al., 2019b] [Simões, 2020].

This line of work started with the use of Machine Learning techniques in the identification of the opponent team and the classification of the opponent robotic soccer formation in the context of the simulation 2D league. The experimental tests performed, using four distinct large datasets of games of the simulation 2D league, generating using 10 different formations for the FCPortugal team enabled us to conclude that SVMs and Neural Networks can identify, with precision, the formations used by the team.

However, improving the performance of a (robo)soccer team through the use of machine learning using game statistics is not trivial. For this task we started by tracking the players throughout the game and develop a high-level automated statistical framework to derive significant information from the game (including passes, shots, ball possession and several other significant events and statistics from the game).

After analysing the robotic soccer games and extracting game events and statistical information we tried to understand which are the game statistics that most influence the final game results. For this, two feature selection algorithms were used (MARS and RreliefF) and a large set of games of the 2D RoboCup competition were chosen. At the end, as the MARS algorithm presented better results, it was used in the next phase. The FC Portugal 2D team with a configurable strategy (namely in the formations and setplays used) was used and three opponents were chosen according to their final position in the previous RoboCup 2D competitions (the best, a middle team and the worst team). Using our game statistic tool, we could calculate the final game statistics through many simulated games and analyse the statistics that most influence the final game results (using only the MARS algorithm). A clustering algorithm was used to group the data and a classifier that can predict the group that better characterize a given input was used (including Support Vector Machines, Neural Networks, Bagging and Random Forests). Finally, the expected best strategy for a group of opponents with similar behaviours according to the maximum of the scored goals was defined including the selected team formations and set plays.

The final online phase consisted in predicting the group for which the given data is expected to be more similar with. The model used for prediction was trained in the previous phase. Finally, a certain team strategy is assigned for a particular way of playing of the opponent. For that, we used the predicted group (obtained in the previous step) and the best strategy per cluster (obtained in the previous phase) in order to obtain the strategy that optimizes the difference of goals scored for a given runtime input data. The final results proved that the using this approach we could slightly increase the FC Portugal performance with all the opponents used in comparison with the normal teat strategy.

This approach, although interesting, clearly showed that just changing the formations and setplays used during the game from a predefined set of formations and setplays, is not enough for significantly improve the team performance. In a soccer match, a cooperative behaviour emerges from the combined execution of simple actions by players. A cooperative behaviour can be planned if players are previously committed to its execution prior to its start or unplanned otherwise. The ability to reproduce some of these behaviours can be useful to help a team achieve better performances.

Thus, we devised an approach to identify, extract setplays from real games, and that lead to a goal while ball possession is kept, and then adapt them in real time to the opponent team. The representation of these behaviours is abstracted using our setplay definition language to promote their reusability. A set of game log files generated with the FC Portugal 2D simulated soccer competition were analysed. The results achieved showed that it is possible to extract and adapt setplays for their reuse. Finally, we developed an approach based on Reinforcement Learning that allows the use of experience to devise the better course of action in set plays with multiple choices. Simulations results showed an improvement on the setplays used and on the team global behaviour.

We are now aiming at learning high-level robotic soccer strategies from scratch through reinforcement learning in the context of the 3D Simulation league and already have very promising results on this line of work.

## 6.5 Novel Stochastic Search Algorithms for Black Box Optimisation

This work was aimed at creating novel stochastic search algorithms for black box optimisation and their application for specific single and multi-agent humanoid learning problems in the context of the RoboCup simulation 3d league. The content of the section is based on the work that led, among others, to the following publications: [Abdolmaleki et al., 2015a], [Abdolmaleki et al., 2015b], [Abdolmaleki et al., 2015c], [Abdolmaleki et al., 2016a], [Abdolmaleki et al., 2016b], [Abdolmaleki et al., 2016c], [Abdolmaleki et al., 2016d], [Abdolmaleki et al., 2016e], [Abdolmaleki et al., 2016f] [Abdolmaleki et al., 2017a], [Abdolmaleki et al., 2017b], [Abdolmaleki et al., 2017c], [Abdolmaleki, 2018], [Abdolmaleki et al., 2019].

Many important applied problems involve finding the best way to accomplish some task. Often this involves finding the maximum or minimum value of some function: the minimum cost for doing a task, minimum time to make a trip, the maximum distance a robot can kick, among many others. Many of these problems can be solved by finding the appropriate function and then using the best techniques to find the maximum or the minimum value required. The machine learning and optimization communities study algorithms that aim to find the best solutions to some criteria for these types of complex problems.

In this work we focus on the subfield of continuous blackbox optimization that uses stochastic search algorithms to find optimal solutions. We have created a collection of new, state-of-the-art, robust, and powerful, stochastic search algorithms, to tackle continuous black box optimisation problems. Stochastic search algorithms aim to repeat a type of mutations that lead to the best solution in a population of candidate solutions. We can model those mutations by a stochastic distribution and, typically, the stochastic distribution is modelled as a multivariate Gaussian distribution. The key idea is to iteratively change the distribution parameters towards better solutions and higher expected fitness. We showed how plain maximisation of the fitness expectation, without bounding the change of the distribution, is destined to fail because of overfitting and the resulting premature convergence. Therefore, we leverage information theoretic trust regions to limit the change of the distribution. Following these guidelines, we introduced two novel general-purpose stochastic search algorithms for black box optimisation named: Trust Region Covariance Matrix Evolution Strategy (TR-CMA-ES) and Model Based Relative Entropy Stochastic Search (MORE).

Being derived from first principles, our proposed approaches can be elegantly extended to contextual learning settings, which allows for learning context dependent stochastic distributions that generates optimal solution for a given context. In other words, we can optimise for multiple related contexts at once and exploit the correlations between related contexts. However, the search distribution typically uses a parametric model that is linear in some hand-defined context features. Finding good context features is a challenging task and so non-parametric methods are often preferred over their parametric counterparts. As a result, we further propose a non-parametric contextual stochastic search algorithm that can learn a non-parametric search distribution for multiple tasks simultaneously.

Our algorithms were applied to a multitude of problems and several variations of these algorithms were used for optimizing all the low-level skills of the FC Portugal team, including the omnidirectional walk, fast kick, contextual kick, controlled kick and get up. The algorithms developed by our team were immediately extended by Google Deepmind and are now state of the art algorithms there… [Abdolmaleki et al. 2018a], [Abdolmaleki et al. 2018b].

## 6.6 Optimized Omnidirectional Walking Skill

Our work on Humanoid Robot Walking was aimed at creating optimized omnidirectional walking skills in the context of the RoboCup simulation 3d league. The content of the section is based on the work that led, among others, to the following publications: [Picado et al., 2009], [Monteiro et al., 2012], [Shafii et al., 2011a], [Shafii et al., 2011b], [Shafii et al., 2013], [Ferreira et al., 2013], [Shafii et al., 2014a], [Abdolmaleki et al., 2014], [Shafii et al., 2014b], [Shafii et al., 2015a], [Shafii et al., 2015b], [Shafii, 2015], [Abdolmaleki et al., 2016e], [Abdolmaleki et al., 2016f], [Abdolmaleki, 2018].

Humanoid robot is a very hard task. Due to the huge controller design space as well as inherently nonlinear dynamics of walking, controlling the gait and the balance of humanoid walking is a very challenging task. There is a large number of approaches to generate humanoid walking, which are mainly divided into model-based and model-free approaches. However, there is still not a clear methodology to address the issue of generating an optimized omnidirectional walking in the sense of creating a fast and energy efficient walk.

In this work we started by developing a very simple walking mechanism and optimized it by using metaheuristics with emphasis for genetic algorithms. Then, we developed biped locomotion approaches in order to generate an optimized omnidirectional walking.

Model-free approaches based on Truncated Fourier Series (TFS) and model-based approaches based on Zero Moment Point (ZMP) were studied to generate optimized walking. The TFS based approach was improved to generate three-dimensional walking motions using all legs degrees of freedom. This improvement enables the robot to perform turn in-place and to walk faster when compared with the previously presented TFS based approaches. The walk was then optimized using state of the art optimization algorithms and latter with our TR-CMA-ES and MORE algorithms for speed and stability.

A ZMP-based approach was developed which can model an omnidirectional walking with variable hip height. The hip height movement is generated using Fourier series and Central Pattern Generators (CPGs) that had been used in model-free walking approaches. Gait optimization approaches are applied to the proposed ZMP-based walk engine with variable hip. This enables the robot to walk faster and also to walk with higher energy efficiency in comparison with the walking generated using the regular cart-table model. The proposed model-based and model-free walk approaches were implemented and tested on both simulated and real NAO humanoid robots. The results showed that by combining and using the techniques developed in model-free and model-based approaches, the robot could perform an optimized omnidirectional walking, concerning energy efficiency and with higher speed.

The proposed optimized walk engine approach was also used in our humanoid soccer robot team that competes in RoboCup international competitions. Our results were encouraging given the fact that the robot performed very well, being able to walk fast and stable in any direction with results very similar, in comparison, with the best RoboCup 3D soccer simulation teams, which use the same simulator, but with a lot more human-like walking behaviour.

## 6.7 Robotic Multi-Agent Learning

This work was aimed at creating robotic multi-agent learning capabilities and its application to several simple domains and also to the Humanoid RoboCup simulation 3d league. The content of the section is based on the work that led, among others, to the following publications: [Simões et al., 2017], [Simões et al., 2018a], [Simões et al., 2018b], [Simões et al., 2018c], [Duarte et al., 2019], [Simões et al., 2019a], [Simões et al., 2019b], [Simões, 2020], [Simões et al., 2020a], [Simões et al., 2020b], [Simões et al., 2020c], [Simões et al., 2020d].

The ability for an agent to coordinate with others within a system is a valuable property in multi-agent systems. Agents either cooperate in the context of a team to accomplish a common goal or compete with opponents to complete selfish goals. Research has shown that learning multi-agent coordination is significantly more complex than learning single agent policies and requires a variety of techniques to deal with the properties of a system where we have a multi-agent learning.

The field of Multi-Agent Reinforcement Learning (MARL) has witnessed a large growth in recent years, with many novel algorithms and techniques surfacing to tackle its challenges. This has been partly due to the re-emergence of deep learning as a solution to handling complex environments, which in turn increases the applications and motivations of developing general MARL algorithms. By using a neural network as a non-linear approximator, algorithms are shown to handle high-dimensional state-spaces and achieve successful policies in complex single- and multi-agent environments. However, they no longer have theoretical proofs of convergence, and only a few algorithms take advantage of the properties of a MAS to learn agent policies.

We developed work to determine how can machine learning be used to achieve coordination within a multi-agent system and how to tackle the increased complexity of such systems and their credit assignment challenges, how to achieve coordination, and how to use communication to improve the behaviour of a team.

Based on our previous research, agents can use function approximators to reduce the environment's complexity, share relevant information with team members, converge to equilibrium policies, and benefit from a centralized learning phase. Results show that the mixture of these techniques yields results that outperform current state-of-the-art approaches. Using deep learning allows agents to approximate continuous high-dimensional state spaces and generalize to new unforeseen situations. Communication protocols can be learned such that agents share relevant information to improve each other's results and compensate for local partial observations of the environment. Equilibrium policies can be found in competitive environments where both opponents can no longer improve, but also cannot be taken advantage of. A centralized learning phase increases the robustness and speed with which a team converges to successful policies, by aggregating observations from all agents and possibly additional information which is not commonly available during execution. Besides several other simpler scenarios, the work was applied on the learning of simples setplays consisting of multi-robot pass and kick scenarios with very promising results.

## 6.8 Humanoid Kick with Controlled Distance

This work was aimed at learning a flexible humanoid robot kick controller, that could be applicable for multiple contexts, such as different kick distances, initial robot position with respect to the ball or both and its application in the RoboCup simulation 3d league. The content of the section is based on the work that led, among others, to the following publications: [Rei et al., 2011], [Cruz et al., 2012], [Rei et al., 2012], [Ferreira et al., 2012a], [Ferreira et al., 2012b], [Abdolmaleki et al., 2016e], [Teixeira et al., 2020], [Abreu et al., 2021], [Abdolmaleki, 2018] [Simões, 2020]

Designing optimal controllers for robotic systems is one of the major tasks in the robotics research field. Hence, it is desirable to have a controller that can control the robot for different tasks or contexts in real time, for example a soccer robot should be able to kick the ball for any desired kick distance which can be chosen from a continuous range of kick distances. We define a task as a context, e.g., a vector of variables that do not change during a task's execution but might change from task to task.

The kick task is one of the most important skills in the context of robotic soccer. Typically, the kick controllers are only applicable for a discretized number of desired distances. For example, three sets of parameters for the kick controller are obtained which are applicable for long, mid, and short distance kicks. Such a controller limits the robot to properly pass to its teammates. Controlling the robot to kick the ball (near)optimally for different distances, allows the agents have a lot more control and options regarding their next decision, which could affect the game's outcome. Our goal is to find a parametric function that given a desired kick distance, outputs the (near) optimal controller parameters. In the other word we would like to obtain a policy that sets the parameters of the robot kick controller given a context s which is the desired kick distance. In order to optimize the robot controller parameters given an objective function, there are many algorithms proposed by the scientific community. However, many of these algorithms usually optimize a parameter set for a single context, such as optimizing a kick for the longest distance or the highest accuracy [Depinet et al., 2014]. In other words, these algorithms fail to generalize the optimized movement for a context to different contexts. In order to generalize the kick motion to, for example, different kicking distances, typically the parameters are optimized for several target contexts independently.

Thus, we researched the learning of a flexible humanoid robot kick controller, i.e., the controller should be applicable for multiple contexts, such as different kick distances, initial robot position with respect to the ball or both. As said before, current approaches typically tune or optimise the parameters of the biped kick controller for a single context, such as a kick with longest distance or a kick for a specific distance. Hence our research objective was to create a flexible kick controller that controls the robot (near) optimally for a continuous range of kick distances and/or initial elative positions of the ball. The goal was to find a parametric function that given a desired kick distance, outputs the (near) optimal controller parameters. We achieve the desired flexibility of the controller by applying a contextual policy search method so that we may generalize the robot kick controller for different distances, where the desired distance is described by a real-valued vector. Since a linear function fails to properly generalize the kick controller over desired kick distances, the optimal parameters of the kick controller are composed by a non-linear function of the desired distances. The kick model used was the long kick initially developed in 2012 by this document author.

For the optimization task we proposed the contextual relative entropy policy search (CREPS) algorithm for optimizing the controller parameters and generalizing them simultaneously and therefore the correlation between different contexts can be exploited in order to accelerate the optimisation. We also combine the update rules of CREPS and CMA-ES resulting to the contextual relative entropy policy search with covariance matrix adaptation (CREPS-CMA) in order to avoid premature convergence.

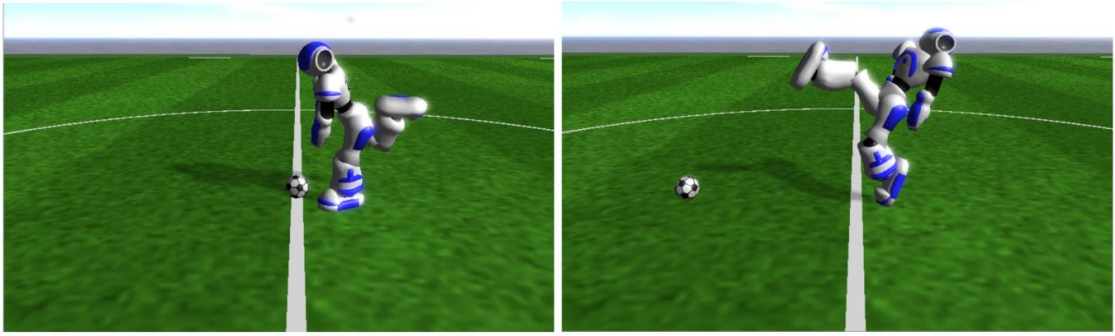

*Figure 9: The initial (left) and final (right) positions of the kick movement*

We used the NAO simulated robot and CREPS-CMA to train the robot by optimising the kick controller. The desired kick distance varies from 2:5m to 12:5m. Figure 9 shows the initial and final position of our typical long kick.

We initialized the search distribution with the hand tuned kick policy, which was able to kick the ball over 15m. We optimized the kick with 1000 iterations. Each iteration generates 20 new samples where the contexts were sampled uniformly- Each sample was evaluated 5 times and was averaged to smooth out the noisy returns. In order to simulate competition conditions, for evaluating each sample, we placed the robot in 5 different positions around the ball and it had to perceive the ball, move towards it, position itself in place and then kick it towards the target goal using the kick controller. The average error achieved was 0:34±0:11m which may be considered an excellent precision given that the simulator has very realistic perception and action errors.

## 6.9 Humanoid Sprinting and Running Skills

This work was aimed at learning a humanoid robot very fast sprinting and running skill and its application in the RoboCup simulation 3d league. The content of the section is based on the work that led, among others, to the following publications: [Abreu et al., 2019a] [Abreu et al., 2019b].

Automated learning of walking skills from scratch is becoming a desirable technique, given the available machine learning tools and the ever-increasing computational speed. Reinforcement learning approaches can obtain state-of-the-art results, while employing model-free algorithms without prior knowledge of the task at hands. This method is particularly useful in the RoboCup 3D Soccer Simulation League, for which the optimization process can be directly accomplished in SimSpark. One of the most relevant low-level skills in this league is concerned with the robot's locomotion. The major flaw found in most team approaches is the lake of linear speed, followed by the lake of capability for rotational speed. So, we believe that dominating these issues could give a serious advantage over the opponent and developed very fast run and sprint behaviours.

Our team developed two main skills: Flexible Run and Fast Sprint. The first one allows the robot to turn to any direction while running. It is composed of a main action and two subtasks which allow the robot to stop or progressively shift to walking. The second skill is more focused on speed and less on turning, and, due to its flexible nature, it can end with a ball kick, in addition to stopping or shifting to walking. Both behaviours control all the NAO's joints except for the head.

The behaviour was learned using an adapted Proximal Policy Optimization (PPO) strategy – a model-free reinforcement learning algorithm. The optimization was performed for 200M time steps using the SimSpark simulator. Table 2 shows relevant statistics for the most successful robot types. The displayed values for the main skills were averaged over 1000 episodes of 10 seconds each.

*Table 2. Sprint and Run Initial Results*

| Skill | Avg. & Max. linear speed along $x$ | Max. rot. speed | Subtask | Duration |
|---|---|---|---|---|
| Sprint | 2.48m/s & 2.62m/s$ | $10°/s$ | Walk Transition | $0.9s$ |
|  |  |  | Stop | $[1, 1.8]s$ |
| Run | $1.41m/s$ & $1.52m/s$ | $160°/s$ | Walk Transition | $0.9s$ |
|  |  |  | Stop | $[1, 1.6]s$ |
|  |  |  | Kick | N.A. |

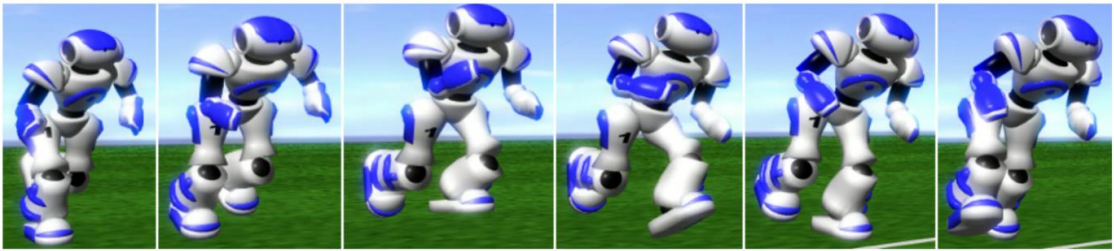

*Figure. 10. Sprint Behaviour*

The fastest sprinter (type 2 robot) achieved an average linear speed of 2.48m/s from the initial standing position until the episodes end (Figure 10). The top speed stabilizes at around 2.62m/s. The best performing runner (type 4) is able to run in a straight line up to 1.52m/s and then rotate to any direction at a maximum of 160◦/s. Sprinting and running can transition into walking smoothly in 45 time steps (equivalent to 0.9s). To bring the robot to a stationary position, it takes between 1s and 1.8s, depending on the gait phase. Fig. 1 shows a sequence of frames taken from the sprinting behaviour at different cycles of the running gait. The robots motion follows a human-like pattern, and actively uses its arms to stabilize itself.

These behaviours were afterwards improved, and the sprint behaviour was improved to achieve more than 3.5 meters per second in the final version. This is more than 3 times the maximum speed achieved by the walking algorithms of the best simulation 3D teams.

Although the sprint and run behaviours are excellent contributions from a scientific viewpoint, they are very difficult to use in normal game play at RoboCup. Thus, in the new FC Portugal code, developed on Python, we opted by not prioritizing the inclusion of these behaviours and prioritize the development of a stable omni-directional walking behaviour, a fast and powerful kicking behaviour, and a fast and flexible dribbling behaviour.

## 6.10 Humanoid Robot Soccer Dribbling Skill

This work was aimed at learning a humanoid robot very fast and reliable dribbling skill and its application in the RoboCup simulation 3d league. The content of the section is based on Kasaei et al., 2021a] [Kasaei et al., 2021b] and on the work that was presented at the RoboCup Simulation 3D Free/Scientific Challenge and that won the corresponding competition, July 2022: [Abreu, Reis and Lau, 2022]. Papers in international conferences and journals about this work will be submitted and published in a very near future.

A humanoid robotic soccer team is only as good as its ability to retain possession and push the ball forward. The former concept requires close control of the ball, while the latter requires a blend of speed, agility, and high-level strategy. Passing allows fast progression to the detriment of close control, while dribbling works in the opposite way. Up to the moment of the development of this work, the league had originated great kicking skills, that are both precise and powerful, being able to kick the ball beyond 18m. However, dribbling skills were still lacking in terms of speed and manoeuvrability. Considering this challenge, we proposed a novel close control ball dribbling behaviour.

The model architecture consists of three main components: a linear inverted pendulum (LIP) model, a shallow neural network, and a predictive controller. The LIP model is an analytical structure that allows the robot to walk in place. Its parameterization adapts the behaviour to different step durations and other gait properties, such as the height of the swing leg, swing span, and stride width.

A shallow neural network is used to learn residual dynamics, i.e., the difference between the walk-in-place behaviour and the dribble skill. The network is composed of only 1 hidden layer with 64 neurons. Its inputs are the state of the walk-in-place behaviour and the robot, as well as the relative position of the ball. The outputs are the residuals of the relative position of feet and hands, which are later converted into joint target angles using inverse kinematics. Finally, these targets are fed to a 1-step predictive controller that takes into consideration the last action sent to the server. The optimization was performed by the Proximal Policy Optimization Reinforcement Learning algorithm, augmented by our Proximal Symmetry Loss [Kasaei et al., 2021a] [Kasaei et al., 2021b] to leverage the robot's symmetry in the sagittal plane.

The dribble can achieve a maximum speed of 1.3 m/s, while keeping the ball at less than 10cm from the midpoint between both feet. For a 90 degree rotation, it has an average turning radius of 0.31m immediately after starting the dribble, and 0.59m if starting to turn at maximum speed. Figure 11 shows the close control dribble, starting with a sharp left turn.

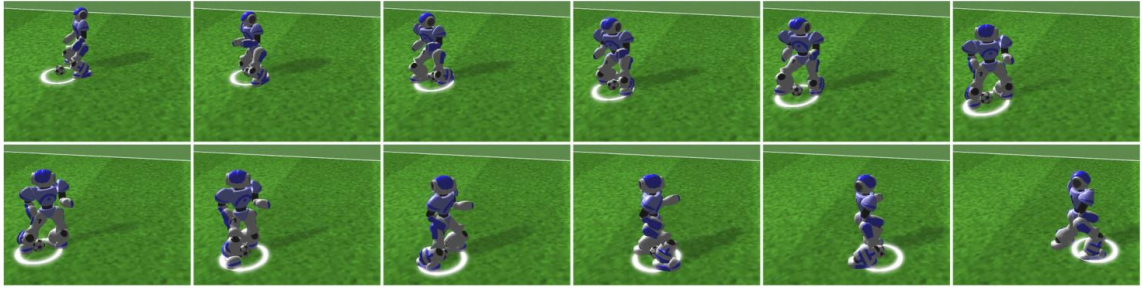

*Figure 11. Close control dribble, starting with a sharp left turn*

This skill was used in RoboCup 2022 but since we considered it unstoppable by the current state of the art teams and we wanted to play an interesting RoboCup competition we used it with very specific conditions. In the first rounds, the dribble was only used in defence to take the ball to the middle field. In the third and final round, it was used only in defence or, in the second half, if the team was not winning the game. In the semi-finals and final it was used only if the team was not winning the game by three or more goals of difference.

## 6.11 Application to Other Leagues

Throughout the year, the work developed was also applied in other RoboCup leagues such as the middle size league [Moreira et al., 2002], [Scolari et al., 2006], [Lau et al., 2009] where all the tactical framework, formations, setplays among many other methodologies were applied. We have also applied the strategical framework to the Simulation Rescue League [Reinaldo et al., 2005] [Certo et al., 2006], [Reis et al., 2006] [Certo et al., 2007a] [Certo et al., 2007b] [Certo et al., 2007c] [Lau et al., 2008] [Alves et al., 2008] [Abdolmaleki et al., 2011] [Abdolmaleki et al., 2012]. Finally, we have also applied our work to the 4-Legged League [Afonso et al., 2004] [Reis, 2006], [Reis et al., 2009] and the Standard Platform League, [Domingues et al., 2011], [Miranda et al., 2013] where, the code used to control the real NAO was exactly the same as the code used for the simulated NAO.

## 7. Conclusions

Coordinating a team of agents, in particular learning individual and collective skills and actions is a complex and non-trivial process. Many different approaches have been proposed in several application domains related to the use of MAS and MRS and how to coordinate and learn in these domains. However, most of the approaches are limited to a single specific domain and do not hold in other application domains. Also, only a few of these models proposed the use high level coordination and deep reinforcement learning for learning individual and cooperative skills.

Throughout the years, our research on robotic soccer was mainly targeted to the development of coordination and machine learning methodologies that could be applied not only to a single robotic soccer problem but to most of the RoboCup leagues and also to other robotic problems that share some similarities. This strategy has limited FCPortugal from winning a large number of competitions but, sacrificing wining, in order to develop excellent quality scientific research work. However, the team won several competitions in different leagues and mostly won many scientific awards at RoboCup. In total, the team won more than 40 awards in international competitions.

In what respects recent awards, the team won the Humanoid Simulation 3D League at RoboCup 2022 competition (Bangkok, Thailand, July 11-17, 2022) scoring a total of 84 goals and conceding only 2, winning all the games except for a draw in the preparation/seeding round. In the semi-finals, FCPortugal won 5-0 against the Chinese team Apollo and, in the Final, the team won 6-1 against the German team magmaOffenburg. FC Portugal also won all the other challenges, including the Proxy challenge and the Free/Scientific Challenge at RoboCup 2022. FC Portugal victory was the first Portuguese victory in a RoboCup main league after CAMBADA in 2008. Other Portuguese victories are limited to FC Portugal victories in 2000, 2002 and 2006.

In what concerns scientific publications, the FC Portugal project and the works briefly described in this lesson summary enabled us to publish more than 100 papers in indexed international conferences and journals, including a very large number of papers in reputed top conferences (CORE A*/A) such as IJCAI, NIPS, GECCO, IROS, IJCNN among many others and reputed journals such as the Journal of Intelligent & Robotic Systems (Springer), Mechatronics (Elsevier), Data and Knowledge Engineering (Elsevier), Soft Computing (Springer), Neurocomputing (Elsevier), Robotics and Autonomous Systems (Elsevier), among many others.

# Acknowledgments

The author would like to acknowledge the contributions of all the colleagues and PhD students that worked and developed research with me in these areas. In particular I would like to thank to Nuno Lau, Eugénio Oliveira, Armando Sousa, Brígida Mónica Faria, António Paulo Moreira, Luís Mota, Fernando Almeida, João Fabro, Nima Shaffi, Abbas Abdolmaleki, David Simões, Miguel Abreu, Tiago Silva and many others.

I would also like to acknowledge the contribution from a large number of FEUP and UAveiro MSc students and other researchers that collaborated with me and with this project. So, the author would also like to thank all the colleagues that collaborated with these projects, giving smart ideas, developing the codes, writing the papers, participating in the competitions and many other tasks, during the last years. Students and colleagues' feedback, discussions, were very useful and, therefore, I would like to thank all the projects students and colleagues for their valuable contributions and feedback.

Although this work is based mostly on the works concerning RoboCup and robotic soccer and thus mostly mentions the work of only 6 of my PhD students, I would also like to thank all my other 17 PhD students that also developed and finished PhD thesis and excellent works of high quality in other areas of Artificial Intelligence, Machine Learning and Intelligent Robotics and to my ongoing 12 PhD students that are also developing excellent work on several AI, ML and Robotics related areas.

Finally, I would like to thank my family and friends, with emphasis to my wife Mónica, but also to my children Clara and Sara, my mother Elsa, my younger brother Miguel, among many others. This year was quite difficult for me with my severe health problems, difficulties on moving the arms, working with the left arm, my surgery and many others. Only with the help of my family and friends I could overrun these issues and deliver this document and the other documents needed for this process. Thank you all!